\documentclass[conference]{IEEEtran}
\usepackage{subfigure}
\usepackage{listings}
\usepackage{xcolor}
\usepackage{amsmath}
\usepackage{pifont}
\usepackage{graphicx}
\usepackage{amssymb}
\usepackage{cleveref}
\usepackage{soul}
\usepackage[shortlabels]{enumitem}
\usepackage{url}
\usepackage{listings}
\usepackage{color}
\usepackage{tikz}
\usepackage{balance}
\usepackage{multirow}
\usepackage{comment}
\usepackage{amsmath}
\usepackage{graphicx}
\usepackage{wrapfig,lipsum,booktabs}
\usepackage{titlesec}
\usepackage[frozencache,cachedir=.]{minted}
\usepackage[symbol]{footmisc}
\usepackage{tablefootnote}

\usepackage{dblfloatfix}

\definecolor{codegreen}{rgb}{0,0.6,0}
\definecolor{codegray}{rgb}{0.5,0.5,0.5}
\definecolor{codepurple}{rgb}{0.58,0,0.82}
\definecolor{backcolour}{rgb}{0.95,0.95,0.92}

\makeatletter
\def\blfootnote{\xdef\@thefnmark{}\@footnotetext}
\makeatother

\lstdefinestyle{mystyle}{
  backgroundcolor=\color{backcolour},   commentstyle=\color{codegreen},
  keywordstyle=\color{magenta},
  numberstyle=\tiny\color{codegray},
  stringstyle=\color{codepurple},
  basicstyle=\ttfamily\footnotesize,
  breakatwhitespace=false,         
  breaklines=true,                 
  captionpos=b,                    
  keepspaces=true,                 
  numbers=left,                    
  numbersep=5pt,                  
  showspaces=false,                
  showstringspaces=false,
  showtabs=false,                  
  tabsize=2,
}

\AtBeginDocument{%
  \providecommand\BibTeX{{%
    \normalfont B\kern-0.5em{\scshape i\kern-0.25em b}\kern-0.8em\TeX}}}

\IEEEoverridecommandlockouts
\begin{document}

\title{BlackMamba: Mixture of Experts for State-Space Models}

\author{Quentin Anthony$^* \quad$ Yury Tokpanov$^* \quad$ Paolo Glorioso$^* \quad$  Beren Millidge$^*$ \\
{
\small
\{quentin, yury, paolo, beren\}@zyphra.com
}\\
{}\\
{
 Zyphra
 \small
}\\
{
\small
 Palo Alto, CA
}
}

\maketitle
\def\thefootnote{*}\footnotetext{All authors contributed equally to this work}\def\thefootnote{\arabic{footnote}}

\setcounter{page}{1}

\begin{abstract}
State-space models (SSMs) have recently demonstrated competitive performance to transformers at large-scale language modeling benchmarks while achieving linear time and memory complexity as a function of sequence length. Mamba, a recently released SSM model, shows impressive performance in both language modeling and long sequence processing tasks. Simultaneously, mixture-of-expert (MoE) models have shown remarkable performance while significantly reducing the compute and latency costs of inference at the expense of a larger memory footprint. In this paper, we present BlackMamba, a novel architecture that combines the Mamba SSM with MoE to obtain the benefits of both. We demonstrate that BlackMamba performs competitively against both Mamba and transformer baselines, and outperforms in inference and training FLOPs. We fully train and open-source 340M/1.5B and 630M/2.8B BlackMamba models on 300B tokens of a custom dataset. We show that BlackMamba inherits and combines both of the benefits of SSM and MoE architectures, combining linear-complexity generation from SSM with cheap and fast inference from MoE. We release all weights, checkpoints, and inference code open-source. \footnote{Inference code at: \url{https://github.com/Zyphra/BlackMamba}}
\end{abstract}


\section{Introduction}
\label{sec:intro}

The advent of Large Language Models (LLMs) built from decoder-only transformer models \cite{bahdanau2014neural,vaswani2017attention} have revolutionized Natural Language Processing (NLP) \cite{radford2019language,brown2020language,touvron2023llama}, along with diverse deep learning application domains such as image processing \cite{dosovitskiy2020image}, time-series \cite{rasul2023lag}, and reinforcement learning \cite{reed2022generalist}. Despite the strong performance and scalability of the transformer architecture, however, there remain significant shortcomings. While maximally expressive, the attention mechanism is computationally demanding both during training and inference, naively requiring both quadratic FLOPs and memory in the sequence length. This limits the context length of transformer models, makes autoregressive generation increasingly expensive with scale, and generally inhibits truly unlimited sequence processing and learning from continual datastreams.

In order to ameliorate these problems, significant effort has recently been directed towards architectural alternatives to the canonical dense attention transformer model. Some of the most promising candidate architectures are State Space Models (SSMs) \cite{gu2023mamba, peng2023rwkv} and Mixture of Experts (MoE) \cite{fedus2022switch,rajbhandari2022deepspeed, jiang2024mixtral}. The key practical benefit of SSMs over transformers is their linear computational complexity with respect to input sequence length (as opposed to the quadratic complexity of transformers). This theoretically enables SSMs to process vastly longer sequences than transformers for a given FLOP budget, and to render autoregressive generation constant in compute without a KV cache. Notable recent examples of SSMs include Mamba \cite{gu2023mamba}, RWKV \cite{peng2023rwkv}, and RetNet \cite{sun2307retentive}, all of which demonstrate efficient long-sequence training and inference, efficient implementations in CUDA, and competitive language modeling task performance to transformers with similar scaling properties. At the same time mixture of expert (MoE) architectures \cite{lepikhin2020gshard,fedus2022review, fedus2022switch,rajbhandari2022deepspeed} have become an emerging advance over dense transformers which allow for significantly reduced training and inference FLOPs required to achieve comparable quality to a comparable dense model. MoE models allow for only a sparse subset of the total parameters to be activated on a single forward pass, relying on a routing function to gate which 'experts' are utilized or not depending on the context. This sparsity decouples the inference cost and parameter count of a model, enabling significantly stronger performance for a given inference budget at the cost of many more parameters and a correspondingly greater memory footprint. 

These architectural improvements over transformers are compelling on their own, but we believe that their combination is a natural next step that could enable significantly improved language modelling speed and performance against the canonical transformer. Specifically, we expect a Mamba-MoE architecture would have the following improvements over a dense transformer:
\newline
\begin{itemize}
\itemsep 1em 
    \item \emph{Mamba}: Linear computational complexity with respect to input sequence length for both training and inference. Autoregressive generation in constant time and memory.
    
    \item \emph{MoE}: Inference latency and training FLOPs of the equivalent smaller dense base model, while preserving model quality close to an equi-parameter dense model.\newline

\end{itemize}

\begin{figure*}[htbp]
  \begin{center}
      \mbox {
          \hspace{-1\columnsep}
          \subfigure[Transformer]
          {
            \includegraphics[width=.092\linewidth]{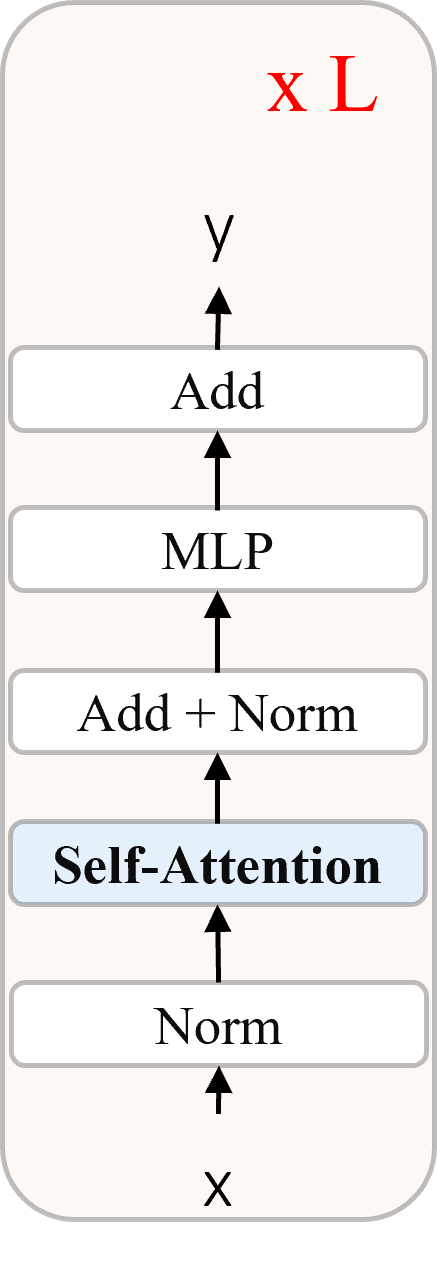}
            \label{fig:transformer-arch}
          }
          \subfigure[Mamba]
          {
            \includegraphics[width=.092\linewidth]{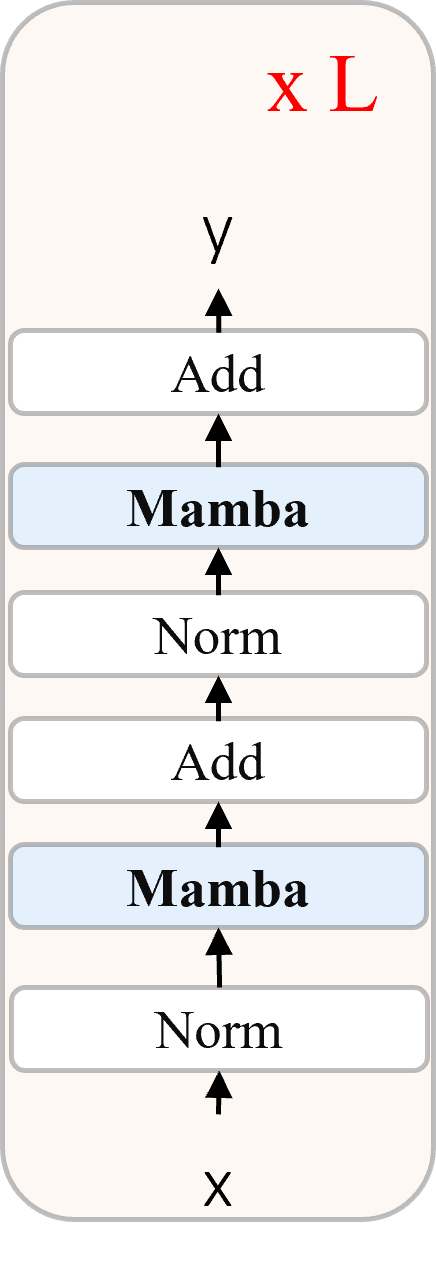}
            \label{fig:mamba-arch}
          }
          \subfigure[Transformer-MoE]
          {
            \includegraphics[width=.35\linewidth]{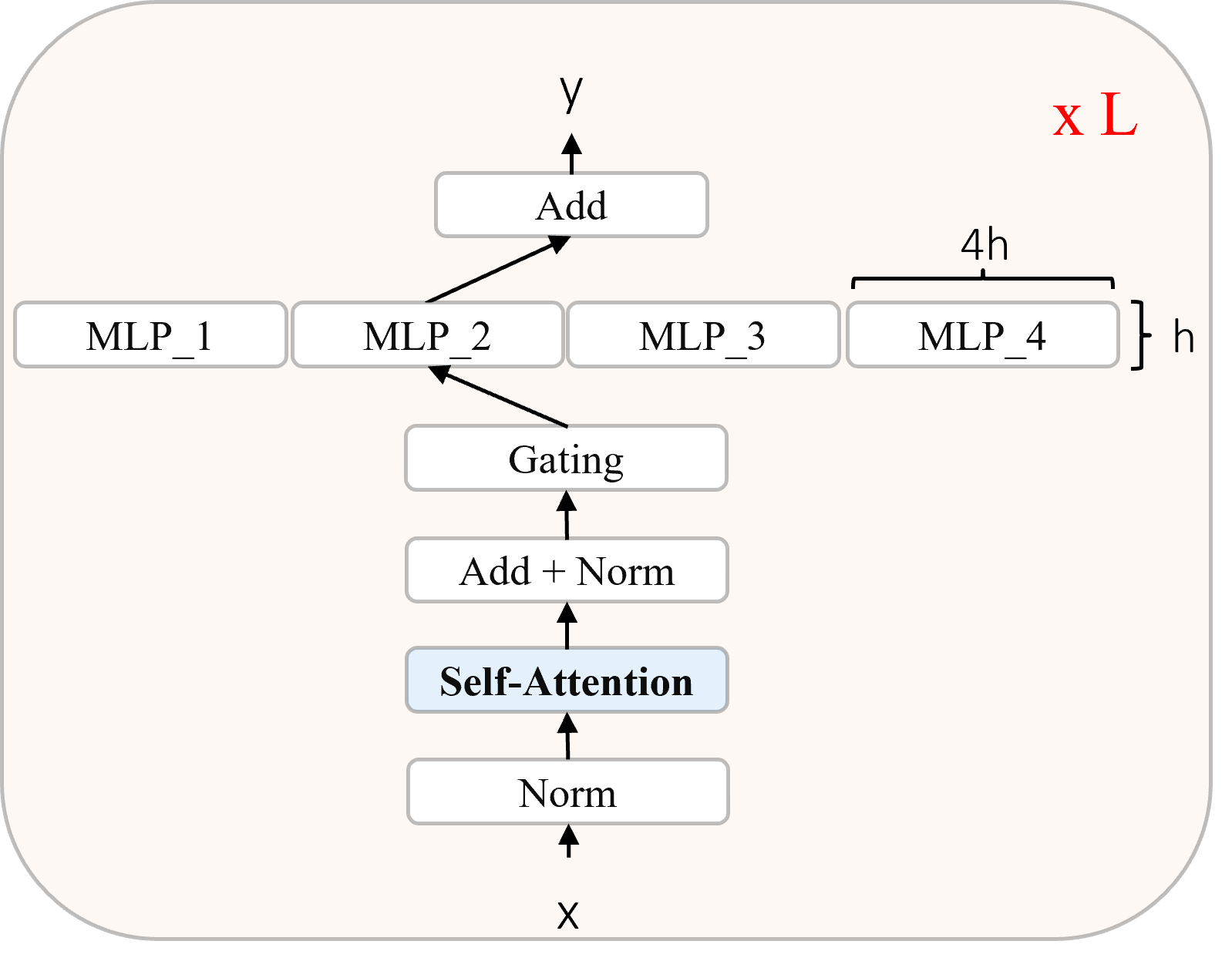}
            \label{fig:transformer-moe-arch}
          }
          \subfigure[Mamba-MoE]
          {
            \includegraphics[width=.35\linewidth]{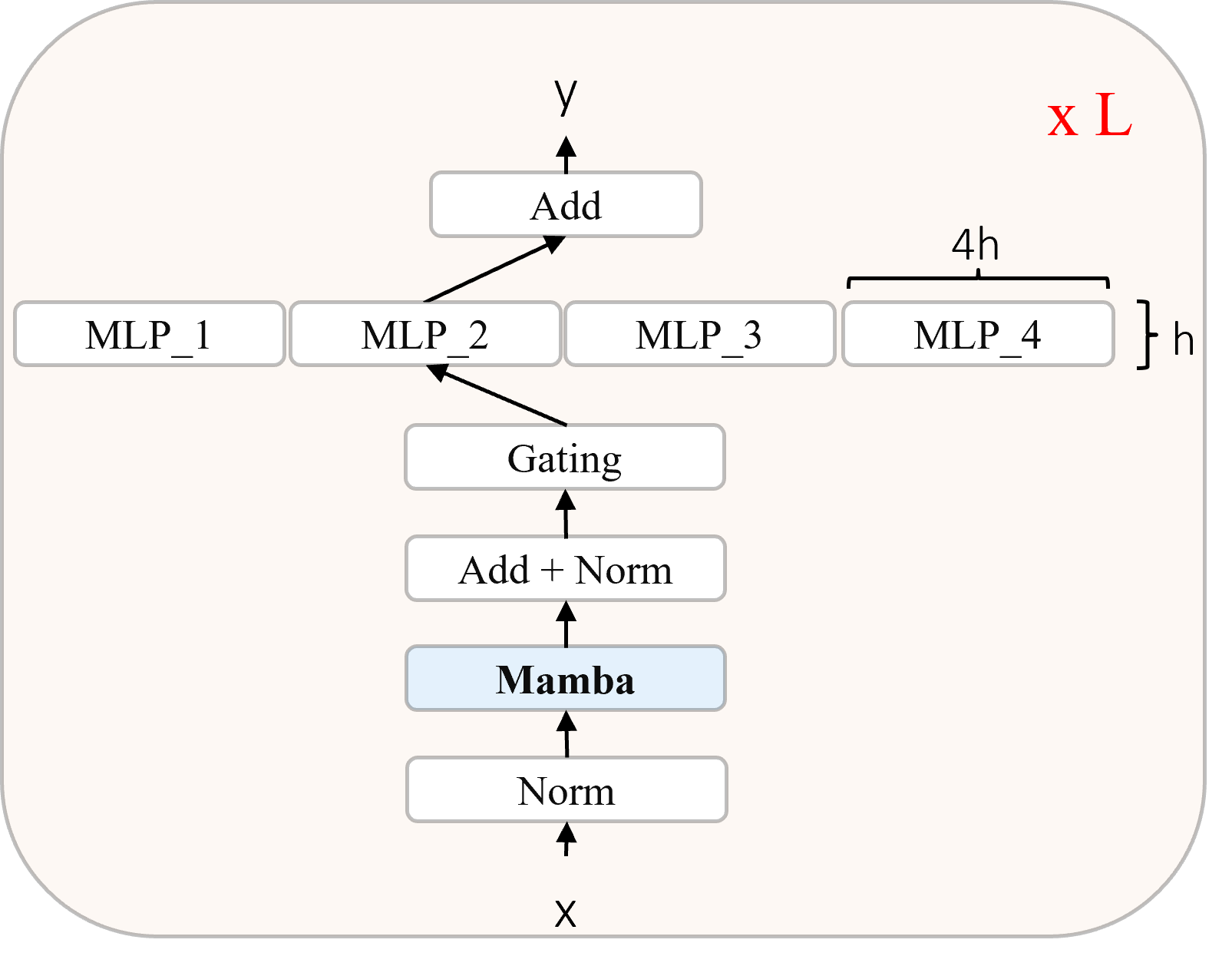}
            \label{fig:mamba-moe-arch}
          }
      }
      \vspace*{-0.5\baselineskip}
      \caption{Architecture of dense transformer, dense Mamba, transformer-MoE, and Mamba-MoE}
      \label{fig:arches}
  \end{center}
\vspace*{-1\baselineskip}
\end{figure*}

\begin{figure*}[htbp]
\centering
    \includegraphics[width=.9\linewidth]{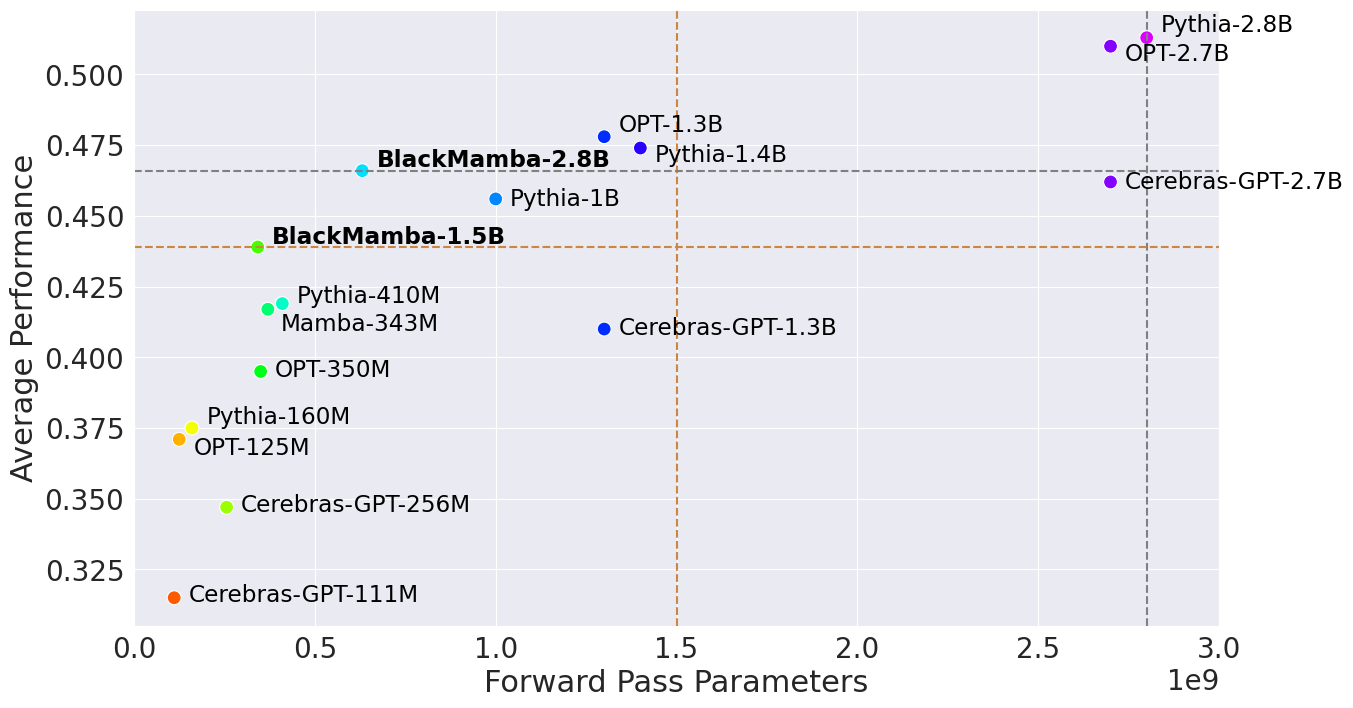}
    \caption{Comparison of BlackMamba average evaluation performance across activated forward parameters.}
    \label{fig:fwd-perf}
\end{figure*}

\begin{figure}[htbp]
\centering
    \includegraphics[width=.99\columnwidth]{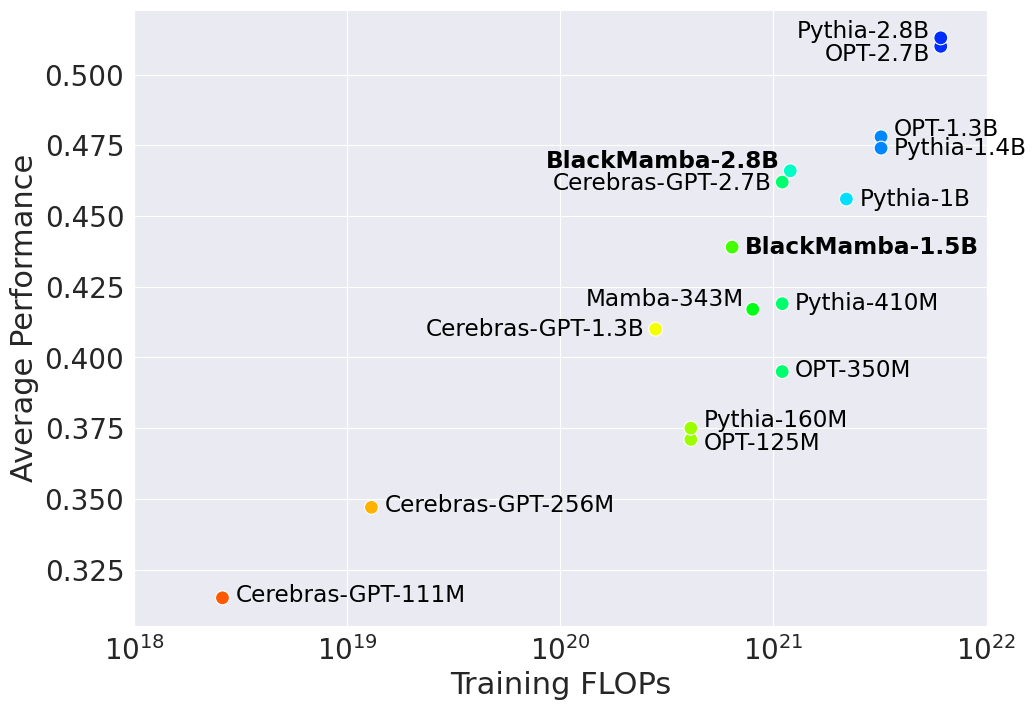}
    \caption{Comparison of BlackMamba average evaluation performance  across training FLOPs.}
    \label{fig:flops-perf}
\end{figure}

In this paper, we begin to demonstrate that these improvements are achievable and that, when put together, these two approaches synergize to produce a model with compelling evaluation performance (Figs. \ref{fig:openbookqa}-\ref{fig:lambada}), compute (Fig. \ref{fig:flops-perf}), and latency advantages (Figs. \ref{fig:generation_latency} and \ref{fig:fwd-perf}) over existing transformer models and which can be trained at a fraction of the FLOP cost for similar performance (Fig. \ref{fig:flops-perf}). We study the MoE routing statistics exhibited by our model across training time and across model depth. Additionally, we introduce a novel initialization for our routing Sinkhorn algorithm which significantly reduces the number of iterations required until convergence, thus improving routing speed.

\section{Contributions}
\label{sec:contributions}
The main achievements of this work are:
\newline
\begin{itemize}
\itemsep 1em 
    \item We design, implement, and evaluate \textbf{BlackMamba}: a combination of alternating attention-free Mamba blocks and routed MLPs.
    \item We train and open-source two BlackMamba Models: 340M/1.5B BlackMamba and 630M/2.8B BlackMamba\footnote{In this paper, we denote an MoE model with $X$ forward-pass parameters and $Y$ total parameters as $X/Y$.}.
    \item We demonstrate that BlackMamba requires significantly fewer training FLOPs to achieve comparable downstream task performance to a dense transformer model.
    \item We explore the compounding inference benefits of the combination of attention-free architectures such as Mamba along with routed sparsity architectures such as MoE.
    \newline
\end{itemize}

The rest of this paper is organized as follows. We first provide an overview of related works on SSM, MoE, and SSM with MoE in Section \ref{sec:related}. We then provide background into the underlying concepts behind SSMs and MoE that are necessary to understand our contributions in Section \ref{sec:background}. Our architecture is described in Section \ref{sec:design}, and its training/inference dynamics are explored in Section \ref{sec:results}. Finally, we describe the implications and limitations of our approach in Section \ref{sec:discussion} along with our conclusions from this work in Section \ref{sec:conclusion}. 

The final checkpoints are open-sourced on HuggingFace with Apache 2.0 licensing, and intermediate training checkpoints are available upon request. Inference code is provided at \url{https://github.com/Zyphra/BlackMamba}.
\section{Background}
\label{sec:background}

\subsection{Transformers}

The transformer architecture \cite{vaswani2017attention} has demonstrated exceptionally strong and consistent performance at language modelling, as well as almost all other sequence processing tasks, remaining state-of-the-art and essentially unchanged since its introduction. The core operation of the transformer is self-attention, which performs a quadratic all-to-all comparison of the dot-product similarities between the embeddings of different tokens in a sequence before normalizing it and performing a linear map to an output vector. Mathematically, self-attention can be written as,
\begin{align}
    z = W_V x \sigma(\frac{1}{\sqrt{d}}x W_Q W_K^T x \circ M)
\end{align}
Where $\sigma$ denotes the softmax function, $M$ denotes a binary mask which enforces specific constraints, such as causal masking, on the computation, the superscript $T$ denotes transposition, and $\circ$ denotes element-wise multiplication. The quadratic cost in sequence length is caused by the $x W_Q W_K^T x$ term which computes a $L \times L$ matrix of similarity scores between the embeddings of different tokens where $L$ is the sequence length.

The transformer model consists of a stack  of self-attention blocks interleaved with multi-layer-perceptron (MLP) blocks which consist of a two-layer MLP with a given activation function. A layer of a transformer model can thus be written as,
\begin{align}
    x_{l+1} = x_l + \text{MLP}(\text{LN}(x_l + \text{attention}(\text{LN}(x_l)))) \label{transformer_layer}
\end{align}
Where $\text{LN}$ represents the layernorm operation which is used to normalize the inputs to the attention and MLP blocks.

\subsection{Mamba}
\label{sec:backg_mamba}

State-space models (SSMs) are a class of sequence models that possess linear complexity with respect to the sequence length. SSMs are more closely related to RNN and CNN architectures than the attention mechanism, and draw inspiration from a continuous dynamical system (depicted in Equation \ref{eq:ssm}) mapping a 1-dimensional function or sequence $x(t) \in \mathbb{R} \mapsto y(t) \in \mathbb{R}$ through an implicit latent state $h(t) \in \mathbb{R}^N$:
\begin{equation}
\label{eq:ssm}
\begin{aligned}
    h'(t) = Ah(t) + Bx(t), \, \, \, \, \, y(t) = Ch(t)     
\end{aligned}
\end{equation}
Where the `time' $t$ now represents the sequence position of a token. A linear dynamical system like this can be efficiently computed in parallel via a convolution or associative scan, while the recurrent form presented above can be utilized for rapid generation at inference time. The fundamental innovation of the Mamba architecture is to make the $A$, $B$, and $C$ matrices of the SSM linearly input-dependent. That is, the new dynamics can be written as,
\begin{align}
        h'(t) = A(x(t))h(t) + B(x(t))x(t), \, \, \, \, \, y(t)= C(x(t))h(t) 
\end{align}
Intuitively, this enables the updates to the SSM's recurrent state to selectively depend upon the tokens being processed, with the SSM being able to decide to store or remove specific information from its recurrent state dynamically. This renders the $A$,$B$,$C$ matrices loosely analogous to the $Q$,$K$,$V$ matrices in attention and significantly increases the expressivity of the SSM block and could potentially enable context to persist much longer in the hidden state than otherwise, since it must exponentially decay in a linear dynamical system with fixed weights. Empirically, \cite{gu2021efficiently} found that this closed much of the gap with transformers.

In practical terms, the recurrent nature of SSMs has long prevented their adoption on the reigning highly-parallel AI hardware like GPUs. However, recent implementations of recurrent and state-space models such as Mamba \cite{gu2023mamba} and RWKV \cite{peng2023rwkv} have mapped these operations efficiently to GPU hardware via parallel scan kernels, thus enabling training of such novel architectures with efficiencies approaching that of well-optimized transformer models.

For more details on Mamba, please see Appendix \ref{sec:appendix-mamba-block} which describes in details the internal computations of a Mamba block as well as \cite{gu2023mamba} and its associated codebase.

\subsection{Mixture of Experts}

Mixture of Expert (MoE) models allow for the inference cost and number of parameters of a model to be decoupled by not activating all parameters on the forward pass and instead routing tokens to specific MLP \emph{experts}. Each expert theoretically specializes in a certain kind of input, and the router (a small neural network) learns which expert to route each token to. Theoretically, this enables the model to maintain almost all the expressivity of the parameter-equivalent dense model at significantly fewer FLOPs.

In standard implementations \cite{fedus2022switch}, which we follow in this paper, the router is a linear layer mapping from tokens to expert indices, and each expert is simply a standard transformer MLP. The expert that the token is routed to is chosen as the top-k of the expert probabilities, where $k$ is a hyperparameter of the architecture. Given an input token to the MoE layer $x$, this is mapped through the router to a probability distribution $p_i(x)$, where $i$ labels the experts. Upon selecting the top-$k$ probabilities, the output of the MoE layer $y$ can be expressed, schematically, as,
\begin{equation}\label{eq:moe} y = \sum_{i\in \text{top-}k} c_i E_i(x)\end{equation}
where $E_1,E_2,\dots$ denote the MLP experts,
\begin{align}
    E_i(x) = W_{\text{out}}f(W_{\text{in}}(\text{LN}(x))
\end{align}
where $f$ is the activation function of the MLP, and $c_i$ are coefficients that are often identified with $p_i$, the probability output by the router of choosing a specific expert. The optimal method for training the router is still uncertain since the ``correct'' expert assignment problem is non-differentiable, and MoE models often struggle with training stability and load-balancing between different experts for hardware efficiency. Nevertheless, MoE models have demonstrated the ability to achieve superior performance for a given compute budget over dense transformer models. Lastly, due to complexity of reporting MoE models, where different papers have reported either the forward pass size of the MoE, the total parameters, or both, we here present a consistent convention of denoting MoE models as: $(\text{forward parameters}) / (\text{total parameters})$. For more details on the MoE architecture and its typical implementation, see \cite{fedus2022review}.
\section{Related Work}
\label{sec:related}

\subsection{State-space Models}

The quadratic complexity of transformers in the sequence length has long been recognized as a primary bottleneck to extremely long context reasoning and understanding. While recent work has pioneered the concept of context-length extension \cite{peng2023yarn, chen2023extending} allowing transformers to be trained at a manageable scale and then inferenced successfully at a significantly longer context, the inference cost in terms of both FLOPs and the memory required for the KV cache remains substantial. 

Early state-space models were inspired by linear dynamical systems which can be efficiently computed as a convolution \cite{gu2021efficiently,poli2023hyena} for sequence processing and as a recurrence for efficient autoregressive generation. However, such models were noticeably less expressive and performant than transformers. 

A number of recent works \cite{sun2307retentive,arora2023zoology} has aimed to increase the expressivity of the state-space model by using input-dependent gating, similar to the QKV matrices of attention, while maintaining the fundamentally linear nature of the state-space recursion. This thus enables efficient implementation via convolution or selective-scan to be maintained while substantially closing the gap to transformer performance in practice. Mamba \cite{gu2023mamba} is a recently released state-space model in line with these previous works which demonstrates strong performance comparable to transformers up to the 2.8B scale, as well as promising scaling laws. Mamba uses input-dependent gating of the inputs to the SSM recursion while maintaining efficient computation via customized selective scan kernels. 

\subsection{Mixture of Experts}

MoE models have been demonstrated to achieve significantly higher performance in both training and inference per FLOP than the equivalent dense models \cite{fedus2022switch, rajbhandari2022deepspeed}. Moreover, scaling laws for MoE models have been put forward \cite{clark2022unified} which show that MoE performance improves smoothly with compute, data, and the number of experts being routed to. This latter is especially important since it provides a route to continually increasing the capability of the model while holding the inference cost fixed.

While MoE models hold significant promise, the architecture still retains many drawbacks. Increasing the number of experts increases the parameter count and hence memory cost substantially, while many works report MoE models being less stable and more challenging to train. Moreover, effective methods for training the router are still open, since the decision to route to an expert or not is discrete and cannot be easily backpropagated through. The large memory cost of MoEs relative to their dense counterparts is especially important for users running on relatively low-end GPUs or when the memory size extends beyond that provided by a single GPU necessitating model-parallelism for inference.

Recently, \cite{jiang2024mixtral} released a powerful open source mixture of experts model which performs competitively with Llama 2 70B \cite{touvron2023llama} and close to GPT-3.5 in evaluations while requiring only the forward pass FLOP cost of the original Mistral 7B model \cite{jiang2023mistral}, thus demonstrating and solidifying the promise of MoE models at scale. The Mixtral architecture also differs in a few ways from earlier MoE work, especially in its use of relatively few experts, a design which we also utilize and have independently found promising for balancing the FLOP and memory cost of MoE models successfully.

\subsection{State-space models with Mixture of Experts}

While both state-space models and Mixture of Experts have been proposed as promising architectures able to improve the computational cost of inferencing language models, no works have ever tested their combination at scale.

Concurrently with this work, \cite{pioro2024moe} demonstrate the performance of extremely small mamba-MoE models in the hundred-million scale of total parameters and the forward pass FLOPs of a 25M model, trained on \textless 10B tokens. In contrast, we demonstrate empirically the scaling potential and performance of such models at meaningful scales in terms of both parameters and data, by training multi-billion parameter models on 300B tokens. Our work thus demonstrates the strong scaling potential of the combination of state-space models and MoE models while resulting in competitive and usable language models which are extremely efficient for inference.
\section{Design}
\label{sec:design}

\subsection{Architecture}
A standard transformer model \cite{vaswani2017attention} consists of interleaved $\text{attention}$ and $\text{MLP}$ blocks added in sequence along a residual stream. The equation for a single transformer layer is written in Equation \ref{transformer_layer}. 

Most MoE architectures simply replace the MLP blocks with a routed expert layer. Our BlackMamba architecture simply replaces both the MLP layer in a transformer with an expert layer, and the attention layer with a mamba SSM layer (see Figure \ref{fig:arches}). A single block of our architecture can thus be written as,
\begin{align}
    x_{l+1} =& x_l + \text{MoE}(\text{LN}(x_l+\text{mamba}(\text{LN}(x_l))))\label{eq:mambaseq}
\end{align}

We trained BlackMamba 340M/1.5B and 630M/2.8B models for 300B tokens on our custom dataset. We used the SwiGLU activation function \cite{shazeer2020glu} for the expert MLPs. We trained with 8 experts, a number that we found balanced well the trade-off between the inference cost and memory footprint of the model. We tested whether sequential or parallel \cite{wang2021gpt} blocks performed  better and found a slight advantage for sequential.  Following \cite{touvron2023llama}, we trained without biases. For the expert router, we used top-1 routing with a Sinkhorn routing function to load-balance between experts. We utilized a novel custom version of the Sinkhorn algorithm which converges substantially faster than vanilla Sinkhorn (Appendix \ref{sec:appendix-sinkhorn}). We trained using the Megatron-LM \cite{shoeybi2019megatron} distributed training framework. The model was trained in bf16 precision. All further model architectures and training hyperparameters are described in Appendix \ref{sec:appendix-model-hparams} and \ref{sec:appendix-training-hparams}, respectively.

\subsection{Dataset}

\begin{figure}[htbp]
\centering
    \includegraphics[width=.8\linewidth]{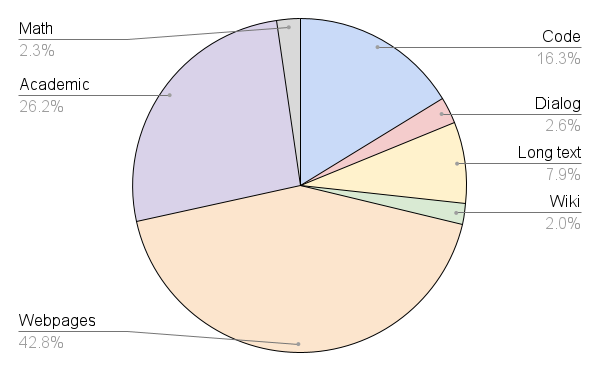}
    \caption{Ratio of data categories in the pretraining dataset of BlackMamba}
    \label{fig:dataset}
\end{figure}

\begin{table}[htbp]
\vspace{1\baselineskip}
\centering
\begin{tabular}{c|c|c}
\hline
\textbf{Dataset} & \textbf{Tokens} & \textbf{Weight} \\
\hline
Pile \cite{gao2020pile} & 300B & 2 \\
SlimPajama \cite{slimpajama} & 600B & 1.2 \\
Starcoder \cite{li2023starcoder} & 250B & 0.75 \\
PeS2o \cite{peS2o} & 50B & 5 \\
Proofpile \cite{azerbayev2023llemma} & 40B & 2 \\
PG19 \cite{rae2019compressive} & 2.2B & 5 \\
\hline
\end{tabular}
\vspace{1ex}
\caption{Dataset subsets and their respective weights in our training mixture}
\label{data_table}
\end{table}

To train BlackMamba, we constructed a custom dataset comprised of a mixture of existing open-source datasets. The subsets included: The Pile \cite{gao2020pile}, SlimPajama \cite{slimpajama}, Starcoder \cite{li2023starcoder}, PeS2o \cite{peS2o}, and ProofPile \cite{azerbayev2023llemma}. The weights for each dataset is provided in Table \ref{data_table}. Tokens were sampled without replacement from each of the subsets according to the probability of sampling from a subset upweighted by these weights. The total dataset comprised 1.8 trillion tokens and thus we trained for significantly less than a single epoch. Preliminary experiments\footnote{We believe that such experiments are not yet rigorous enough for publication, and will be included in future work.} show that long-form text and academic work appears to improve natural language modeling when included in the pretraining phase, so we weigh it heavily in the training recipe. Further, we find that including significant portions of code and math during the pretraining phase meaningfully improves the model's reasoning ability.  We note that this dataset is comparatively heavy on unfiltered web data and contains many duplicates due to the upweighting of smaller subsets, which may limit the quality of the model and leaves significant room for improvement, as well as potentially causing undue memorization of specific common fragments.
\section{Results}
\label{sec:results}

\begin{figure}[htbp]
\centering
    \includegraphics[width=1\linewidth]{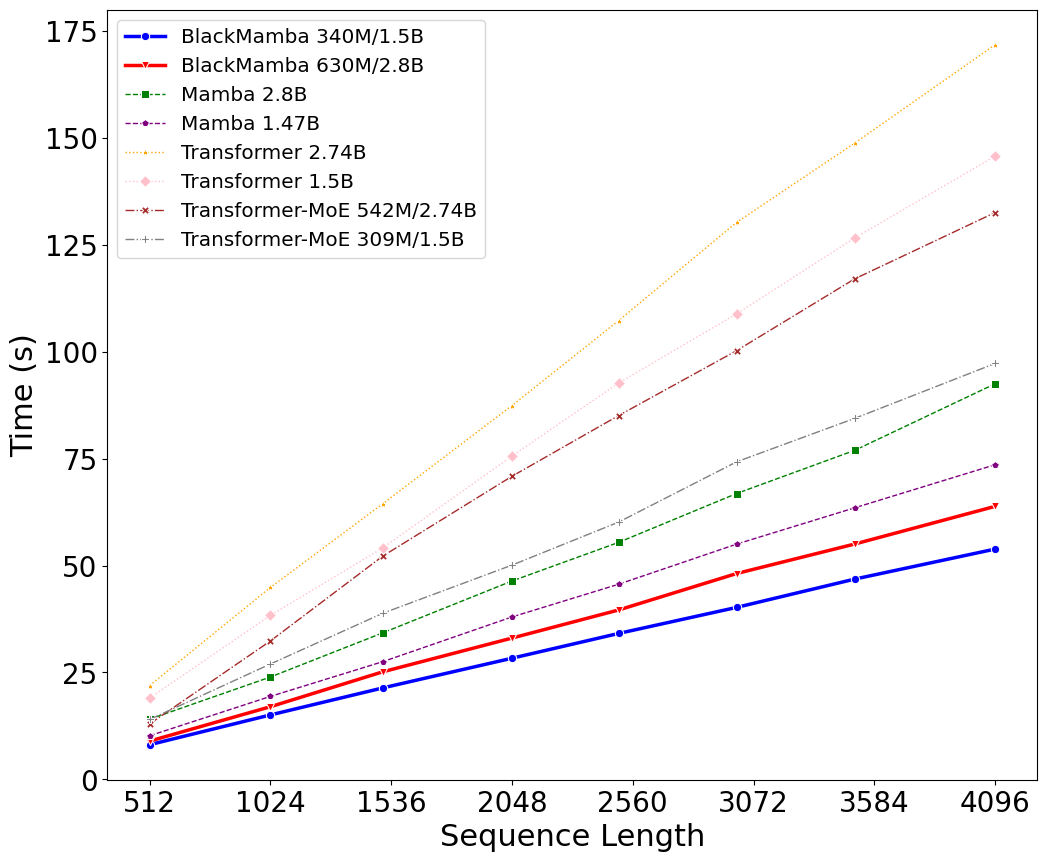}
    \caption{Generation latency of BlackMamba compared to dense transformers, dense mamba, and transformer-MoE}
    \label{fig:generation_latency}
\end{figure}

\begin{figure}[t]
\centering
    \includegraphics[width=1\linewidth]{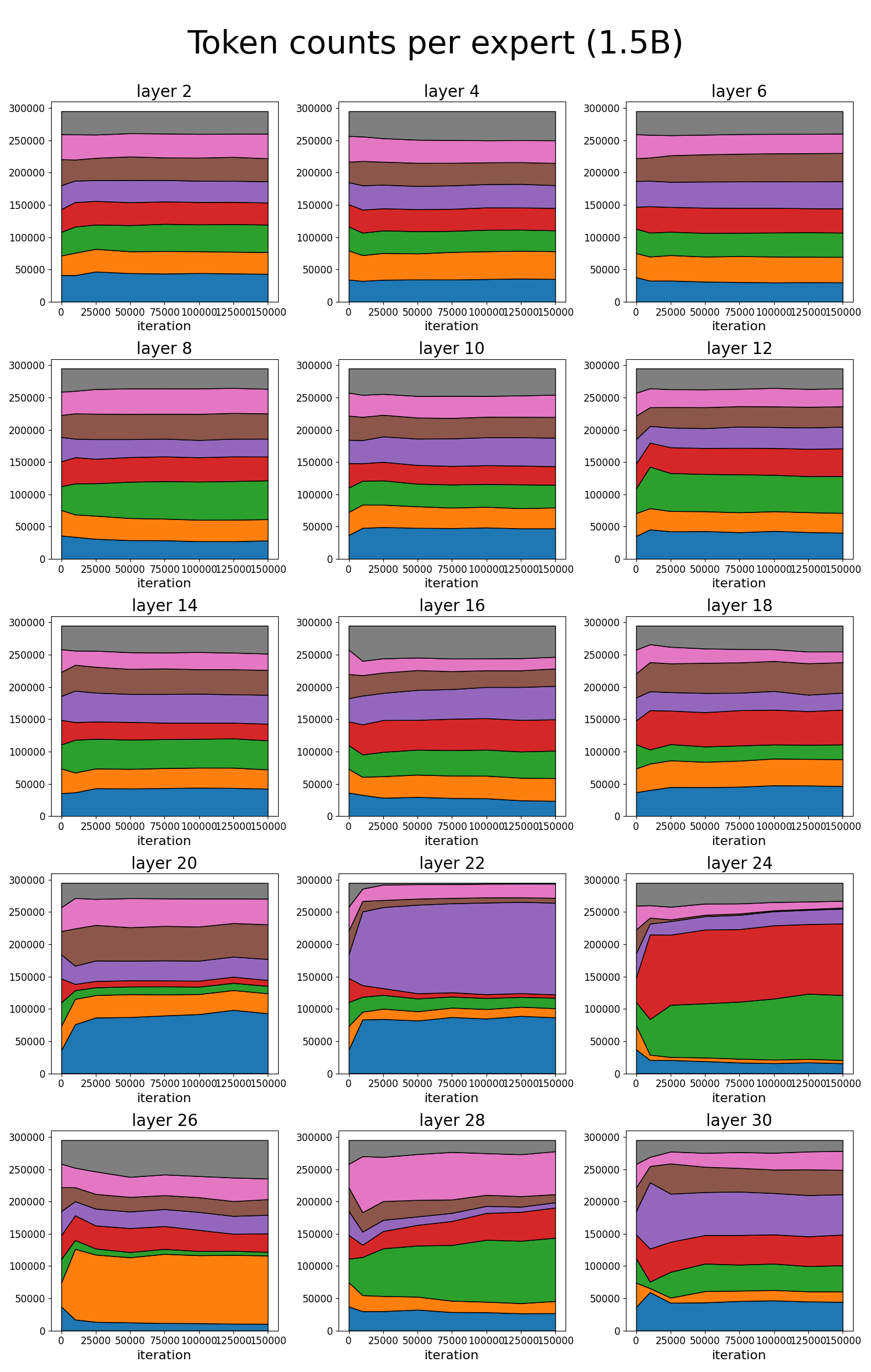}
    \caption{Token distribution across experts in 340M/1.5B BlackMamba}
    \label{fig:token_count_1p5}
\end{figure}

\begin{figure}[t]
\centering
    \includegraphics[width=1\linewidth]{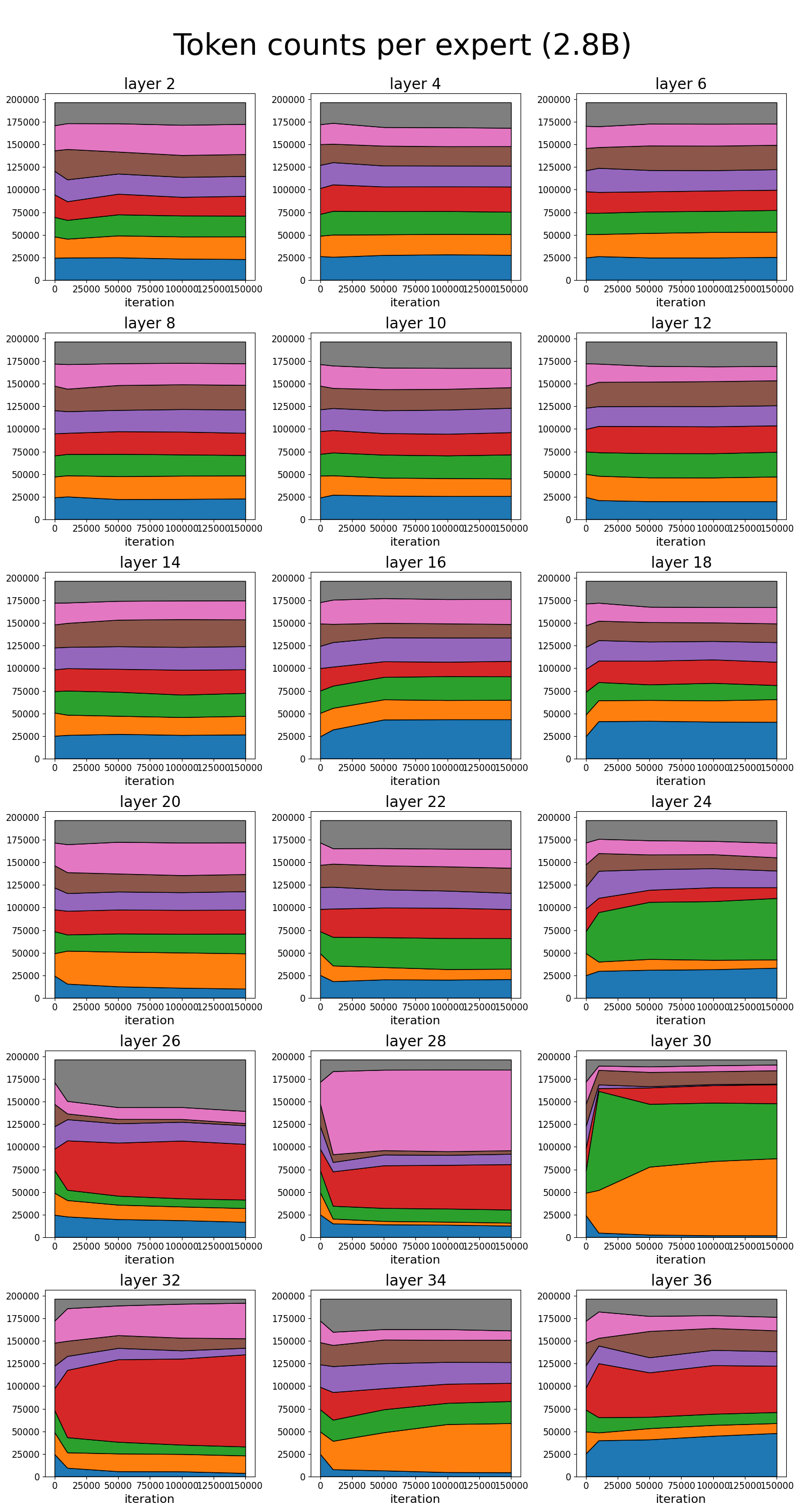}
    \caption{Token distribution across experts in 630M/2.8B BlackMamba}
    \label{fig:token_count_2p8}
\end{figure}

\renewcommand*{\thefootnote}{\fnsymbol{footnote}}
\begin{table*}[b]
\large
\centering
\resizebox{\textwidth}{!}{%
\begin{tabular}{c|ccc|ccccccc|c}
\hline
\textbf{}           & Forward Pass Parameters & Total Parameters & Training FLOPs  & HellaSwag      & PIQA           & WinoGrande     & Lambada        & ARC-e          & ARC-c          & OpenBookQA     & Downstream Average \\
\hline
Cerebras-GPT        & 111M             & 111M                    & 2.6e18          & 0.268\footnote[1]          & 0.594          & 0.488          & 0.194          & 0.38           & 0.166          & 0.118          & 0.315              \\
OPT                 & 125M             & 125M                    & 4.1e20          & 0.313\footnote[1]          & 0.63           & 0.503          & 0.379          & 0.435          & 0.189          & 0.166          & 0.371              \\
Pythia              & 160M             & 160M                    & 4.1e20          & 0.293\footnote[1]          & 0.627          & 0.519          & 0.389          & 0.452          & 0.181          & 0.16           & 0.375              \\
Cerebras-GPT        & 256M             & 256M                    & 1.3e19          & 0.286\footnote[1]          & 0.613          & 0.511          & 0.293          & 0.41           & 0.17           & 0.158          & 0.347   \\ 
\textbf{BlackMamba} & \textbf{342M}    & \textbf{1.5B}           & \textbf{6.4e20} & \textbf{0.365}\footnote[1] & \textbf{0.690} & \textbf{0.526} & \textbf{0.493} & \textbf{0.561} & \textbf{0.241} & \textbf{0.196} & \textbf{0.439}     \\
OPT                 & 350M             & 350M                    & 1.1e21          & 0.366\footnote[1]           & 0.644          & 0.523          & 0.452          & 0.44           & 0.207          & 0.176          & 0.395              \\

Mamba            & 343M             & 343M                    & 8.0e20             & 0.335\footnote[1]         & 0.665         & 0.516          & 0.453          & 0.540          & 0.212           &0.198                &0.417                    \\
Pythia              & 410M             & 410M                    & 1.1e21          & 0.333\footnote[1]          & 0.668          & 0.53           & 0.505          & 0.504          & 0.213          & 0.178          & 0.419              \\
\hline
\textbf{BlackMamba} & \textbf{631M}    & \textbf{2.8B}           & \textbf{1.2e21} & \textbf{0.397}\footnote[1] & \textbf{0.712} & \textbf{0.521} & \textbf{0.542} & \textbf{0.603} & \textbf{0.245} & \textbf{0.242} & \textbf{0.466}  \\
Pythia              & 1B               & 1B                      & 2.2e21          & 0.376\footnote[1]          & 0.705          & 0.545          & 0.566          & 0.559          & 0.243          & 0.196          & 0.456              \\
OPT                 & 1.3B             & 1.3B                    & 3.2e21          & 0.4537\footnote[1]          & 0.717          & 0.595          & 0.579          & 0.57           & 0.234          & 0.234          & 0.478              \\
Cerebras-GPT        & 1.3B             & 1.3B                    & 2.8e20          & 0.384\footnote[1]          & 0.664          & 0.521          & 0.462          & 0.508          & 0.224          & 0.166          & 0.410              \\
Pythia              & 1.4B             & 1.4B                    & 3.2e21          & 0.398\footnote[1]          & 0.711          & 0.565          & 0.604          & 0.576          & 0.256          & 0.204          & 0.474              \\

OPT                 & 2.8B             & 2.8B                    & 6.1e21          & 0.606\footnote[1]          & 0.738          & 0.61           & 0.637          & 0.609          & 0.268          & 0.25           & 0.510              \\
Cerebras-GPT        & 2.8B             & 2.8B                    & 1.1e21          & 0.488\footnote[1]          & 0.701          & 0.559          & 0.567          & 0.571          & 0.246          & 0.206          & 0.462              \\
Pythia              & 2.8B             & 2.8B                    & 6.1e21          & 0.451\footnote[1]          & 0.737          & 0.612          & 0.654          & 0.629          & 0.288          & 0.22           & 0.513              \\
\hline
  
\end{tabular}%
}
\vspace{1\baselineskip}
\caption{Evaluation performance of BlackMamba compared to similar models}
\label{tab:evaluations}
\end{table*}
\renewcommand*{\thefootnote}{\arabic{footnote}}

To ensure a fair comparison vs Mamba, we trained our own 340M Mamba model with the same dataset and training hyperparameters reported for BlackMamba. This Mamba 340M model used a hidden size of 1152 and 34 mamba layers. Notably, BlackMamba performs significantly better than equivalent pretrained models (both transformer and Mamba) for the same forward pass model size at inference time, as well as training FLOPs. 
In Figure \ref{fig:generation_latency}, we plot the time taken to autoregressively generate a sequence of a given length starting from an initial one-token prompt as a function of sequence length. We observe that the established latency benefits of both Mamba and MoE models are combined in BlackMamaba to result in inference times significantly faster than canonical transformer models, MoE transformer models, and pure Mamba models. Moreover, the inference advantage of BlackMamba increases with greater sequence lengths, making BlackMamba extremely competitive at long sequence generation. Moreover, although not reflected in this Figure, it must be recognized that while the transformer inference latency also increases linearly, this is due to KV caching which has additional linearly increasing memory requirements and would eventually OOM on large enough sequences. By contrast, Mamba models (and BlackMamba) can generate sequences of arbitrary length with a constant memory footprint.

Figures \ref{fig:token_count_1p5} and \ref{fig:token_count_2p8} illustrate the token counts assigned to each expert in each layer of the BlackMamba 340M/1.5B and the BlackMamba 630M/2.8B models respectively. Most layers display a high degree of expert balance, as expected by our improved Sinkhorn algorithm. Yet, intriguingly, both models show a clear transition towards expert imbalance in the final layers (at layer 20 for the 340M/1.5B model and layer 25 for the 630M/2.8B model). This may reflect increasing specialization in later layers or else reflect numerical instabilities that develop deeper in the network. While the true cause of this imbalance remains unknown, we also note that a similar pattern of imbalance but convergence to a stable expert assignment has also been observed in previous MoE models \cite{he2022fastermoe}.

\renewcommand*{\thefootnote}{\fnsymbol{footnote}}
In Table \ref{data_table}, we report evaluation scores of BlackMamba against a suite of open-source pretrained language model baselines. We re-evaluated all models on the same version of lm-eval (v0.3.0) that we evaluated our own model on\footnote[1]{We use the non-normalized HellaSwag evaluation results in this paper, which differs from those in \cite{gu2023mamba}}. 
\renewcommand*{\thefootnote}{\arabic{footnote}}

In Appendix \ref{sec:appendix-evals}, we provide evaluation scores for our model during training from checkpoints taken every 10k steps. We generally found relatively smooth but noisy improvements in the evaluation scores during training. To prevent overfitting to the evaluations, we only looked at the evaluation scores after the models had finished training and did not use them for model selection.

Additionally, in Appendix \ref{sec:appendix-sinkhorn}, we describe a novel initialization for the classical Sinkhorn algorithm used for MoE routing which significantly improves convergence speed of the approach, often requiring only a single iteration for convergence. This provides notable speed improvements for the routed expert layers and results in a similar latency to a router with a regularized balancing loss, providing superior balancing performance while requiring much less complexity of implementation.

Finally, in Appendix \ref{sec:appendix-mamba-block}, we provide a detailed mathematical description of the internal computations of a Mamba Block and in Appendix \ref{sec:appendix-params-flops}, we provide detailed and explicit formulas for computing the parameters and training FLOPs for Mamba and MoE models which we hope aid the community in further developing and exploring novel SSM and MoE architectures.

\section{Discussion}
\label{sec:discussion}

This work is a preliminary exploration and validation of the core concept of combining together recent advances in SSMs with MoEs to produce a highly competitive and efficient architecture both in terms of inference and generation time and training FLOPs. While initial results are promising, much work needs to be done to improve both the SSM and MoE components as well as investigation of the optimal way to approach their combination. We ultimately believe that by exploring promising emerging architectures architectures and novel ways of merging and combining them, significant advances in performance, efficiency, and speed can be obtained over standard transformer recipes.

We believe that our work can be extended in many fruitful directions. The evaluations presented in this paper are limited in scope. While we provide general coverage of standard pure language modelling evaluations in the zero-shot setting, the performance of the model in the many-shot in-context-learning setting remains unexplored. Additionally, there are many facets of behaviour of our models which we have not explicitly investigated. We have not tested for factual accuracy, profanity, toxicity, or any other socially undesirable text generation. Similarly, our training dataset blend has not been explicitly scraped for socially undesirable tokens, nor its potential overlap with any evaluation tasks\footnote{In particular, we are aware of the possibility of evaluation dataset contamination present in the widely used RedPajama dataset \cite{elazar2023whats}, and will attempt to explicitly deduplicate this dataset if used in future work.}. Although our dataset remains imperfect, we have released all major details as to its construction and composition with the goal of aiding community understanding of the effects of dataset on pretraining performance and model behaviours.

In terms of scaling laws, while our models are highly competitive for a given inference cost and FLOP training budget, it is impossible to make conclusive scaling extrapolations both in terms of data and parameter counts with only two models trained on 300 billion tokens. Additionally, many of our training hyperparameters may be suboptimal as we performed only basic hyperparameter tuning of the learning rate. Additionally, while we performed some ablations on the core architecture, it is possible that a superior method of combining state-space models and mixture of experts would provide significant benefits. Additionally, the efficacy and performance of well-established finetuning and RLHF pipelines for instruction following and general alignment, as well as standard techniques for parameter-efficient-finetuning of SSM and MoE models remains almost completely unexplored, as does how such models perform under quantization. 

Our work also raises interesting questions as to the modularity of different neural network components that can be placed together into a final model architecture. We show that it is relatively straightforward to combine SSM blocks with MoE blocks from transformers at scale with competitive performance. However, whether Mamba and other SSMs show the same degree of improvement in performance with MoE as transformers remains uncertain, as well as whether combining these architectural pieces has the same effect on the internal representations and behaviours of the model. Additionally, it is unclear the extent to which routing serves the same function in BlackMamba as in more classical transformer MoE models.

\section{Conclusion}
\label{sec:conclusion}

In this paper, we have proposed, implemented and trained BlackMamba, a model that combines both recent advances in state-space models and mixture-of-experts into a single unified architecture. We demonstrate that our BlackMamba architecture performs highly competitively to strong pretrained LLM baselines in terms of inference cost and training flops, and moreover that it inherits the reduced training and generation FLOPs of both SSMs and MoEs simultaneously. Moreover, we show that BlackMamba is capable of rapid generation with both linear time and memory cost. We release BlackMamba 340M/1.5 and 630M/2.8 billion parameter models and intermediate checkpoints, as well as inference code, under a permissive Apache 2.0 license with the goal of enabling and fostering further study, experimentation, and understanding of the potential of this novel architecture by the broader community.

\section*{Acknowledgement}
\noindent The Zyphra team would like to thank Adam Ibrahim for helpful discussions and comments on training stability and hyperparameters, and Albert Gu for general discussions on state space models.

\clearpage

\bibliographystyle{IEEEtran}
\bibliography{main}

\begin{thebibliography}{10}
\providecommand{\url}[1]{#1}
\csname url@samestyle\endcsname
\providecommand{\newblock}{\relax}
\providecommand{\bibinfo}[2]{#2}
\providecommand{\BIBentrySTDinterwordspacing}{\spaceskip=0pt\relax}
\providecommand{\BIBentryALTinterwordstretchfactor}{4}
\providecommand{\BIBentryALTinterwordspacing}{\spaceskip=\fontdimen2\font plus
\BIBentryALTinterwordstretchfactor\fontdimen3\font minus \fontdimen4\font\relax}
\providecommand{\BIBforeignlanguage}[2]{{%
\expandafter\ifx\csname l@#1\endcsname\relax
\typeout{** WARNING: IEEEtran.bst: No hyphenation pattern has been}%
\typeout{** loaded for the language `#1'. Using the pattern for}%
\typeout{** the default language instead.}%
\else
\language=\csname l@#1\endcsname
\fi
#2}}
\providecommand{\BIBdecl}{\relax}
\BIBdecl

\bibitem{bahdanau2014neural}
D.~Bahdanau, K.~Cho, and Y.~Bengio, ``Neural machine translation by jointly learning to align and translate,'' \emph{arXiv preprint arXiv:1409.0473}, 2014.

\bibitem{vaswani2017attention}
A.~Vaswani, N.~Shazeer, N.~Parmar, J.~Uszkoreit, L.~Jones, A.~N. Gomez, {\L}.~Kaiser, and I.~Polosukhin, ``Attention is all you need,'' \emph{Advances in neural information processing systems}, vol.~30, 2017.

\bibitem{radford2019language}
A.~Radford, J.~Wu, R.~Child, D.~Luan, D.~Amodei, I.~Sutskever \emph{et~al.}, ``Language models are unsupervised multitask learners,'' \emph{OpenAI blog}, vol.~1, no.~8, p.~9, 2019.

\bibitem{brown2020language}
T.~Brown, B.~Mann, N.~Ryder, M.~Subbiah, J.~D. Kaplan, P.~Dhariwal, A.~Neelakantan, P.~Shyam, G.~Sastry, A.~Askell \emph{et~al.}, ``Language models are few-shot learners,'' \emph{Advances in neural information processing systems}, vol.~33, pp. 1877--1901, 2020.

\bibitem{touvron2023llama}
H.~Touvron, L.~Martin, K.~Stone, P.~Albert, A.~Almahairi, Y.~Babaei, N.~Bashlykov, S.~Batra, P.~Bhargava, S.~Bhosale \emph{et~al.}, ``Llama 2: Open foundation and fine-tuned chat models,'' \emph{arXiv preprint arXiv:2307.09288}, 2023.

\bibitem{dosovitskiy2020image}
A.~Dosovitskiy, L.~Beyer, A.~Kolesnikov, D.~Weissenborn, X.~Zhai, T.~Unterthiner, M.~Dehghani, M.~Minderer, G.~Heigold, S.~Gelly \emph{et~al.}, ``An image is worth 16x16 words: Transformers for image recognition at scale,'' \emph{arXiv preprint arXiv:2010.11929}, 2020.

\bibitem{rasul2023lag}
K.~Rasul, A.~Ashok, A.~R. Williams, A.~Khorasani, G.~Adamopoulos, R.~Bhagwatkar, M.~Bilo{\v{s}}, H.~Ghonia, N.~V. Hassen, A.~Schneider \emph{et~al.}, ``Lag-llama: Towards foundation models for time series forecasting,'' \emph{arXiv preprint arXiv:2310.08278}, 2023.

\bibitem{reed2022generalist}
S.~Reed, K.~Zolna, E.~Parisotto, S.~G. Colmenarejo, A.~Novikov, G.~Barth-Maron, M.~Gimenez, Y.~Sulsky, J.~Kay, J.~T. Springenberg \emph{et~al.}, ``A generalist agent,'' \emph{arXiv preprint arXiv:2205.06175}, 2022.

\bibitem{gu2023mamba}
A.~Gu and T.~Dao, ``Mamba: Linear-time sequence modeling with selective state spaces,'' \emph{arXiv preprint arXiv:2312.00752}, 2023.

\bibitem{peng2023rwkv}
B.~Peng, E.~Alcaide, Q.~Anthony, A.~Albalak, S.~Arcadinho, H.~Cao, X.~Cheng, M.~Chung, M.~Grella, K.~K. GV \emph{et~al.}, ``Rwkv: Reinventing rnns for the transformer era,'' \emph{arXiv preprint arXiv:2305.13048}, 2023.

\bibitem{fedus2022switch}
W.~Fedus, B.~Zoph, and N.~Shazeer, ``Switch transformers: Scaling to trillion parameter models with simple and efficient sparsity,'' \emph{The Journal of Machine Learning Research}, vol.~23, no.~1, pp. 5232--5270, 2022.

\bibitem{rajbhandari2022deepspeed}
S.~Rajbhandari, C.~Li, Z.~Yao, M.~Zhang, R.~Y. Aminabadi, A.~A. Awan, J.~Rasley, and Y.~He, ``Deepspeed-moe: Advancing mixture-of-experts inference and training to power next-generation ai scale,'' in \emph{International Conference on Machine Learning}.\hskip 1em plus 0.5em minus 0.4em\relax PMLR, 2022, pp. 18\,332--18\,346.

\bibitem{jiang2024mixtral}
A.~Q. Jiang, A.~Sablayrolles, A.~Roux, A.~Mensch, B.~Savary, C.~Bamford, D.~S. Chaplot, D.~d.~l. Casas, E.~B. Hanna, F.~Bressand \emph{et~al.}, ``Mixtral of experts,'' \emph{arXiv preprint arXiv:2401.04088}, 2024.

\bibitem{sun2307retentive}
Y.~Sun, L.~Dong, S.~Huang, S.~Ma, Y.~Xia, J.~Xue, J.~Wang, and F.~Wei, ``Retentive network: A successor to transformer for large language models (2023),'' \emph{URL http://arxiv. org/abs/2307.08621 v1}.

\bibitem{lepikhin2020gshard}
D.~Lepikhin, H.~Lee, Y.~Xu, D.~Chen, O.~Firat, Y.~Huang, M.~Krikun, N.~Shazeer, and Z.~Chen, ``Gshard: Scaling giant models with conditional computation and automatic sharding,'' \emph{arXiv preprint arXiv:2006.16668}, 2020.

\bibitem{fedus2022review}
W.~Fedus, J.~Dean, and B.~Zoph, ``A review of sparse expert models in deep learning,'' \emph{arXiv preprint arXiv:2209.01667}, 2022.

\bibitem{gu2021efficiently}
A.~Gu, K.~Goel, and C.~R{\'e}, ``Efficiently modeling long sequences with structured state spaces,'' \emph{arXiv preprint arXiv:2111.00396}, 2021.

\bibitem{peng2023yarn}
B.~Peng, J.~Quesnelle, H.~Fan, and E.~Shippole, ``Yarn: Efficient context window extension of large language models,'' \emph{arXiv preprint arXiv:2309.00071}, 2023.

\bibitem{chen2023extending}
S.~Chen, S.~Wong, L.~Chen, and Y.~Tian, ``Extending context window of large language models via positional interpolation,'' \emph{arXiv preprint arXiv:2306.15595}, 2023.

\bibitem{poli2023hyena}
M.~Poli, S.~Massaroli, E.~Nguyen, D.~Y. Fu, T.~Dao, S.~Baccus, Y.~Bengio, S.~Ermon, and C.~R{\'e}, ``Hyena hierarchy: Towards larger convolutional language models,'' \emph{arXiv preprint arXiv:2302.10866}, 2023.

\bibitem{arora2023zoology}
S.~Arora, S.~Eyuboglu, A.~Timalsina, I.~Johnson, M.~Poli, J.~Zou, A.~Rudra, and C.~R{\'e}, ``Zoology: Measuring and improving recall in efficient language models,'' \emph{arXiv preprint arXiv:2312.04927}, 2023.

\bibitem{clark2022unified}
A.~Clark, D.~De~Las~Casas, A.~Guy, A.~Mensch, M.~Paganini, J.~Hoffmann, B.~Damoc, B.~Hechtman, T.~Cai, S.~Borgeaud \emph{et~al.}, ``Unified scaling laws for routed language models,'' in \emph{International Conference on Machine Learning}.\hskip 1em plus 0.5em minus 0.4em\relax PMLR, 2022, pp. 4057--4086.

\bibitem{jiang2023mistral}
A.~Q. Jiang, A.~Sablayrolles, A.~Mensch, C.~Bamford, D.~S. Chaplot, D.~d.~l. Casas, F.~Bressand, G.~Lengyel, G.~Lample, L.~Saulnier \emph{et~al.}, ``Mistral 7b,'' \emph{arXiv preprint arXiv:2310.06825}, 2023.

\bibitem{pioro2024moe}
M.~Pi{\'o}ro, K.~Ciebiera, K.~Kr{\'o}l, J.~Ludziejewski, and S.~Jaszczur, ``Moe-mamba: Efficient selective state space models with mixture of experts,'' \emph{arXiv preprint arXiv:2401.04081}, 2024.

\bibitem{shazeer2020glu}
N.~Shazeer, ``Glu variants improve transformer,'' \emph{arXiv preprint arXiv:2002.05202}, 2020.

\bibitem{wang2021gpt}
B.~Wang and A.~Komatsuzaki, ``Gpt-j-6b: A 6 billion parameter autoregressive language model,'' 2021.

\bibitem{shoeybi2019megatron}
M.~Shoeybi, M.~Patwary, R.~Puri, P.~LeGresley, J.~Casper, and B.~Catanzaro, ``Megatron-lm: Training multi-billion parameter language models using model parallelism,'' \emph{arXiv preprint arXiv:1909.08053}, 2019.

\bibitem{gao2020pile}
L.~Gao, S.~Biderman, S.~Black, L.~Golding, T.~Hoppe, C.~Foster, J.~Phang, H.~He, A.~Thite, N.~Nabeshima \emph{et~al.}, ``The pile: An 800gb dataset of diverse text for language modeling,'' \emph{arXiv preprint arXiv:2101.00027}, 2020.

\bibitem{slimpajama}
\BIBentryALTinterwordspacing
D.~Soboleva, F.~Al-Khateeb, R.~Myers, J.~Steeves, J.~Hestness, and N.~Dey, ``Slimpajama: A 627b token cleaned and deduplicated version of redpajama,'' 7 2023. [Online]. Available: \url{https://www.cerebras.net/blog/slimpajama-a-627b-token-cleaned-and-deduplicated-version-of-redpajama}
\BIBentrySTDinterwordspacing

\bibitem{li2023starcoder}
R.~Li, L.~B. Allal, Y.~Zi, N.~Muennighoff, D.~Kocetkov, C.~Mou, M.~Marone, C.~Akiki, J.~Li, J.~Chim \emph{et~al.}, ``Starcoder: may the source be with you!'' \emph{arXiv preprint arXiv:2305.06161}, 2023.

\bibitem{peS2o}
L.~Soldaini and K.~Lo, ``{peS2o (Pretraining Efficiently on S2ORC) Dataset},'' {Allen Institute for AI}, Tech. Rep., 2023, oDC-By, \url{https://github.com/allenai/pes2o}.

\bibitem{azerbayev2023llemma}
Z.~Azerbayev, H.~Schoelkopf, K.~Paster, M.~D. Santos, S.~McAleer, A.~Q. Jiang, J.~Deng, S.~Biderman, and S.~Welleck, ``Llemma: An open language model for mathematics,'' \emph{arXiv preprint arXiv:2310.10631}, 2023.

\bibitem{rae2019compressive}
J.~W. Rae, A.~Potapenko, S.~M. Jayakumar, and T.~P. Lillicrap, ``Compressive transformers for long-range sequence modelling,'' 2019.

\bibitem{he2022fastermoe}
J.~He, J.~Zhai, T.~Antunes, H.~Wang, F.~Luo, S.~Shi, and Q.~Li, ``Fastermoe: modeling and optimizing training of large-scale dynamic pre-trained models,'' in \emph{Proceedings of the 27th ACM SIGPLAN Symposium on Principles and Practice of Parallel Programming}, 2022, pp. 120--134.

\bibitem{elazar2023whats}
Y.~Elazar, A.~Bhagia, I.~Magnusson, A.~Ravichander, D.~Schwenk, A.~Suhr, P.~Walsh, D.~Groeneveld, L.~Soldaini, S.~Singh, H.~Hajishirzi, N.~A. Smith, and J.~Dodge, ``What's in my big data?'' 2023.

\bibitem{eval-harness}
\BIBentryALTinterwordspacing
L.~Gao, J.~Tow, B.~Abbasi, S.~Biderman, S.~Black, A.~DiPofi, C.~Foster, L.~Golding, J.~Hsu, A.~Le~Noac'h, H.~Li, K.~McDonell, N.~Muennighoff, C.~Ociepa, J.~Phang, L.~Reynolds, H.~Schoelkopf, A.~Skowron, L.~Sutawika, E.~Tang, A.~Thite, B.~Wang, K.~Wang, and A.~Zou, ``A framework for few-shot language model evaluation,'' 12 2023. [Online]. Available: \url{https://zenodo.org/records/10256836}
\BIBentrySTDinterwordspacing

\bibitem{zellers2019hellaswag}
R.~Zellers, A.~Holtzman, Y.~Bisk, A.~Farhadi, and Y.~Choi, ``Hellaswag: Can a machine really finish your sentence?'' 2019.

\bibitem{bisk2019piqa}
Y.~Bisk, R.~Zellers, R.~L. Bras, J.~Gao, and Y.~Choi, ``Piqa: Reasoning about physical commonsense in natural language,'' 2019.

\bibitem{sakaguchi2019winogrande}
K.~Sakaguchi, R.~L. Bras, C.~Bhagavatula, and Y.~Choi, ``Winogrande: An adversarial winograd schema challenge at scale,'' 2019.

\bibitem{paperno2016lambada}
D.~Paperno, G.~Kruszewski, A.~Lazaridou, Q.~N. Pham, R.~Bernardi, S.~Pezzelle, M.~Baroni, G.~Boleda, and R.~Fernández, ``The lambada dataset: Word prediction requiring a broad discourse context,'' 2016.

\bibitem{clark2018think}
P.~Clark, I.~Cowhey, O.~Etzioni, T.~Khot, A.~Sabharwal, C.~Schoenick, and O.~Tafjord, ``Think you have solved question answering? try arc, the ai2 reasoning challenge,'' 2018.

\bibitem{mihaylov2018suit}
T.~Mihaylov, P.~Clark, T.~Khot, and A.~Sabharwal, ``Can a suit of armor conduct electricity? a new dataset for open book question answering,'' 2018.

\bibitem{biderman2023pythia}
S.~Biderman, H.~Schoelkopf, Q.~G. Anthony, H.~Bradley, K.~O’Brien, E.~Hallahan, M.~A. Khan, S.~Purohit, U.~S. Prashanth, E.~Raff \emph{et~al.}, ``Pythia: A suite for analyzing large language models across training and scaling,'' in \emph{International Conference on Machine Learning}.\hskip 1em plus 0.5em minus 0.4em\relax PMLR, 2023, pp. 2397--2430.

\bibitem{sinkhorn1967concerning}
R.~Sinkhorn and P.~Knopp, ``Concerning nonnegative matrices and doubly stochastic matrices,'' \emph{Pacific Journal of Mathematics}, vol.~21, no.~2, pp. 343--348, 1967.

\end{thebibliography}

\clearpage
\appendix

\subsection{Model Hyperparameters}
\label{sec:appendix-model-hparams}

\begin{table}[H]
\centering
\begin{tabular}{c|c|c}
\hline
\textbf{Hyperparameter} & \textbf{1.5B} & \textbf{2.8B} \\
\hline
Number of Layers & 30 & 36 \\
Hidden Size & 1152 & 1472 \\
Number of Experts & 8 & 8 \\
Sequence Length & 2048 & 2048 \\
State Size & 16 & 16 \\
Convolution Dimension & 4 & 4 \\
FFN Hidden Size & 3072 & 3872 \\
Expansion Factor & 2 & 2 \\
\hline
\end{tabular}
\vspace{1ex}
\caption{Architecture hyperparameters for the 340M/1.5B and 630M/2.8B models}
\label{arch_haram_table}
\end{table}

\subsection{Training Hyperparameters}
\label{sec:appendix-training-hparams}

\begin{table}[H]
\centering
\begin{tabular}{c|c|c}
\hline
\textbf{Hyperparameter} & \textbf{340M/1.5B} & \textbf{630M/2.8B} \\
\hline
Learning Rate & 0.0002 & 0.00015  \\
Batch Size  & 2064384 tokens & 2162688 tokens  \\
Dropout & 0.0 & 0.0\\
Learning Rate Schedule & cosine & cosine\\
Min Learning Rate & 0.00002 & 0.00002  \\
Weight Decay & 0.0 & 0.0\\
\hline
\end{tabular}
\vspace{1ex}
\caption{Training hyperparameters for the 340M/1.5B and 630M/2.8B models}
\label{training_hparam_table}
\end{table}

\subsection{Mamba Block Internals}
\label{sec:appendix-mamba-block}

In this appendix, we provide a precise and detailed walkthrough of the core computations that comprise a Mamba block. Mamba derives from a line of work on state-space models, which are expressive recurrent models which have recently been shown capable of competing with transformers on large scale sequence modelling. The recurrence of these models enables them to be used efficiently for generation without a KV cache and causes them to scale in FLOPs and memory linearly in the sequence length. The core insight is to utilize recurrence \cite{gu2021efficiently} or selective scan \cite{gu2023mamba} to efficiently map the central recurrence to parallel GPU hardware. The base of all such models is the following state-space equations (in continuous time):

\begin{align}
\frac{dh}{dt} &= A\,h + B\,x \\
y &= C\,h
\end{align}
which define a classical linear time-invariant dynamical system. Here $h$ denotes the state of a system at one instant. $A$ denotes a matrix which governs the 'natural dynamics' of $h$ over time. $x$ denotes a 'control' input to the system -- i.e. one provided by the controller or experimenter and $B$ denotes a dynamics matrix which controls how $x$ interacts with system. Finally, the states are transformed into 'observations', denoted $y$, through the observation matrix denoted $C$.

The Mamba block utilizes this dynamical system across tokens as its core computation implemented as a hardware efficient selective scan. The innovation of Mamba specifically is to make the $A$,$B$,and $C$ matrices a linear function of the input $x$, analogous to the $Q$,$K$,$V$ matrices of a self-attention block. Beyond this, Mamba wraps the SSM component in a linear projection to and from the residual stream and a convolution of the input, as well as an additional gating projection path which gates the output of the SSM based on a projection of the input to the block.

We denote the input to the mamba block $x$, the recurrent hidden state $h$, the sequence length as $l$. We set the hidden recurrent state dimension to some factor of the input dimension.

The mamba block contains matrices $A$ which defines the dynamics for the recurrent state, $B$ which is the projection for the inputs, $C$ which is the projection to the outputs $y$, the matrix $D$ which is a learnable bias on the output, a discretization timestep $dt$, and a gating vector $z$. The Mamba block also performs a linear projection of the input x and z prior to the SSM with weight matrices $W_x$ and $W_z$ and an output projection matrix $W_y$.

The computation inside a Mamba block runs as follows. First, the $x$ and $z$ projections are computed. This projection occurs for every token in the sequence independently.
\begin{align}
 x &= W_x\,x \\
 z &= W_z\,z
\end{align}
Secondly, after the projection, the Mamba block performs a 1d convolution ($\ast$) across the input sequence embeddings. This convolution cannot be merged with the projection $W_x$ because this projection acts at the embedding level, and the convolution is acting at the sequence of tokens level.

\begin{align}
 x_t = W_{filter\_t} \ast x_t
\end{align}

The input-dependent ‘weights’ $B$, $C$, and $dt$ can then be computed, which are analogous to the Query, Key, and Value weights in attention. 

\begin{align}
B &= W_B\,x \\
C &= W_C\,x \\
dt &= W_D\,x
\end{align}

The matrix $A$ is trained with a special initialization given in the matrix below.  Note that updates are trained via the parameterization $\ln(A)$, presumably to make $A$ positive and to improve stability, and then computed as $A = \exp(\,\ln(A)\,)$.
\begin{align}
A &=
	\begin{bmatrix}
	1 & 2 & 3 & \cdots \\
	1 & 2 & 3 & \cdots \\
	\vdots \\
	\end{bmatrix}
\end{align}

The weights are then discretized prior to use in the SSM kernel. Note that the discretization for B does not follow Equation 4 in \cite{gu2023mamba}.
\begin{align}
dt &= \text{softplus}(dt + dt_{\text{bias}}) \\
dA &= \exp(-A \,dt) \\
dB &= B\,dt
\end{align}

A single step of the ssm is then performed to obtain the new recurrent state.  Note that $h^+ \to h$ when $dt \to 0$, as expected  

\begin{align}
h^+ = dA\,h + dB\,x
\end{align}

From the new recurrent state, the output  $C\,h^+$ can be computed.  This output is also gated by the learnt gating vector z and passed through a final output projection before being addded back into the residual stream.
\begin{align}
y &= C\,h^+ + D\,x \\
y &= \text{silu}(z) \, y \\
y &= W_y\,y \\
\end{align}

The output of the SSM block is then the hidden state $h^+$ and the output $y$.

A Mamba block can operate in two modes. The first mode is the recurrent method, which directly follows the steps described here. This approach is linear in both memory and computational cost for a single step since it only utilizes the recurrent state to predict the next token. The second way is to run the SSM across the whole sequence at once using the 'selective scan' operation and kernel introduced by \cite{gu2023mamba}. For further reference on the implementation of the selective scan refer to \cite{gu2023mamba}.

\subsection{Computing Parameters and FLOPs for Mamba-MoE}
\label{sec:appendix-params-flops}

Let us denote the embedding dimension $D$, the Mamba inner state as $I$, the recurrent state dimension $H$, the dt rank $dt$ and the convolution dimension $C$. We denote the batch size $B$ and the sequence length $L$.

The number of parameters in a Mamba block can then be computed as,
\begin{align}
    \underbrace{3ID}_{W_x, W_z, W_y} + 2I(\underbrace{H}_{W_A, W_B} + \underbrace{dt}_{W_{dt}} + \underbrace{\frac{C}{2}}_{\text{conv}}) + \underbrace{I}_{D} + \underbrace{2D}_{\text{layernorm}}
\end{align}

The number of parameters in a MoE block can be computed as
\begin{align}
    \underbrace{8D^2E}_{\text{experts}} + \underbrace{DE}_{\text{router}}
\end{align}
Where $E$ is the number of experts in the layer. For a network of $L$ layers, there are thus $\frac{L}{2}$ Mamba blocks and $\frac{L}{2}$ MoE blocks.

To begin approximating the number of FLOPs involved in a single Mamba block, we make the following observation.

Given two matrices $A \in \mathcal{R}^{K \times M}$ and $B \in \mathcal{R}^{M \times J}$, then the total FLOPs involved in the matrix product $AB$ is approximately $2KMJ$, where the factor of $2$ arises from the fact that matrix multiplication requires both a multiply and an add operation. In the following calculations, we assume that the matrix multiplications dominate the total FLOP count of the model and hence ignore the nonlinearities, layernorms, and other computations. 

First, let us consider the projection operation involving the weights $W_x$,$W_z$, and $W_y$. All are of shape $I \times D$ and hence the total FLOPs for these are $6IDLB$.

There is also the convolution which can be treated as a single $I \times C$ matrix multiply requiring $2ICLB$ FLOPs.

Now, we turn to the SSM block itself. We first compute the input-dependent $B$ and $C$ matrices requiring a matrix multiply of shape $I \times H$ each thus resulting in $4IH$ FLOPs. The $A$ matrix is not multiplied by the input but goes through an elementwise transform costing $IH$ FLOPs. The $dt$ projection first goes through an elementwise operation of order $I$ FLOPs. 

Next, the discretization. The $A$ matrix is multiplied by the $dt$ vector resulting, costing $IH$ FLOPs. The $B$ matrix is multiplied by the input costing $2IH$ FLOPs. The SSM linear state space step itself is just a matrix multiply and add so costs $2IH$ FLOPs, and then the output projection using the $C$ matrix also costs $2IH$ FLOPs. Putting this all together, we obtain the following expression,
\begin{align} BLI(\underbrace{11H}_{W_x, W_z, W_y, \text{SSM}} + \underbrace{4dt}_{\text{dt proj, discretization}} + \underbrace{1}_{\text{dt nonlinearity}}) + \underbrace{IH}_{A}
\end{align}

The MoE blocks consist of $E$ standard mlp blocks and a router. The FLOPs for each mlp block is simply $16D^2$ since there are two weight matrices of shape $4D \times D$, and a multiply and add per matrix multiply. The router cost is simply $2DE$. Putting this together, we obtain $DE(16D +2)$ FLOPs for an MoE block.

\subsection{Evaluations During Training}
\label{sec:appendix-evals}

We evaluate BlackMamba on a suite of eight diverse evaluation tasks in the zero-shot setting. We use the EleutherAI evaluation harness (version 0.3.0) \cite{eval-harness}. Specifically, we evaluate our models on the HellaSwag \cite{zellers2019hellaswag}, PIQA \cite{bisk2019piqa}, WinoGrande \cite{sakaguchi2019winogrande}, Lambada \cite{paperno2016lambada}, ARC \cite{clark2018think} (both the easy and challenge versions), and OpenBookQA \cite{mihaylov2018suit}. The evaluations were run on model checkpoints taken every $10,000$ steps. We observe that most evaluation metrics appear to increase smoothly but noisily throughout training, before appearing to plateau towards their final values. This is broadly in line with previous findings in the Pythia model suite \cite{biderman2023pythia}, which find relatively smooth improvements across training in many of their evaluation metrics. This provides some evidence that the development of capabilities in language models occurs smoothly and can be tracked during training and perhaps predicted ahead of time. Two evaluation metrics, however, WinoGrande and BoolQ, violate this trend for reasons that we do not currently understand. We note that \cite{biderman2023pythia} also observe no consistent trend on Winogrande. Between the BlackMamba $340M/1.5B$ and $630M/2.8B$ models, we observe a clear benefit of scale at the same iteration and token count on most evaluations. In addition, we observe significant noise in some of the evaluation metrics which may suggest that small differences in evaluations between different LLMs may not be significant.

\begin{figure}[H]
\centering
    \includegraphics[width=.8\linewidth]{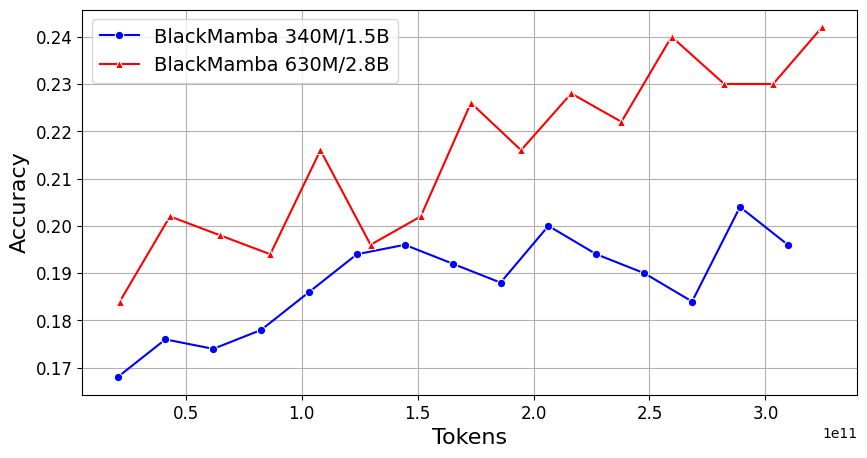}
    \caption{OpenBookQA evaluation accuracy over time}
    \label{fig:openbookqa}
\end{figure}

\begin{figure}[H]
\centering
    \includegraphics[width=.8\linewidth]{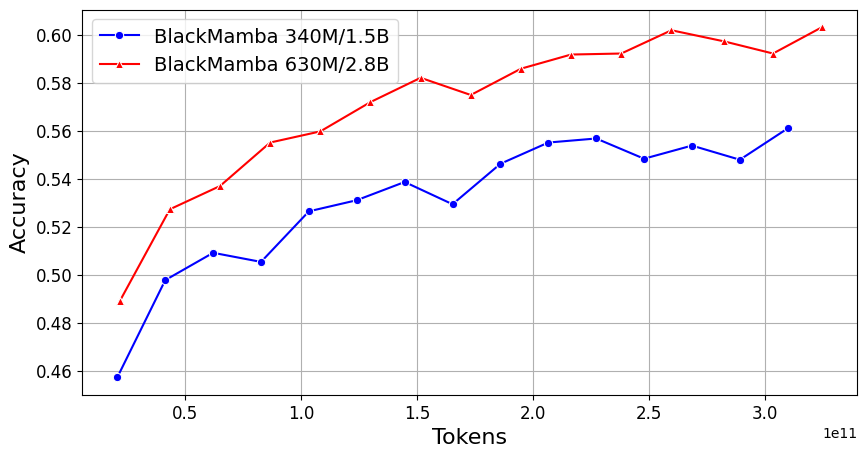}
    \caption{ARC-Easy evaluation accuracy over time}
    \label{fig:arc-easy}
\end{figure}

\begin{figure}[H]
\centering
    \includegraphics[width=.8\linewidth]{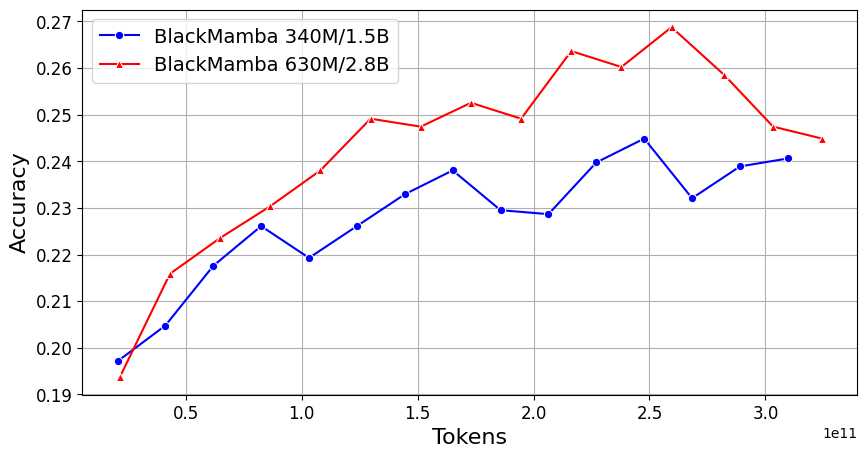}
    \caption{ARC-Challenge evaluation accuracy over time}
    \label{fig:arc-challenge}
\end{figure}

\begin{figure}[H]
\centering
    \includegraphics[width=.8\linewidth]{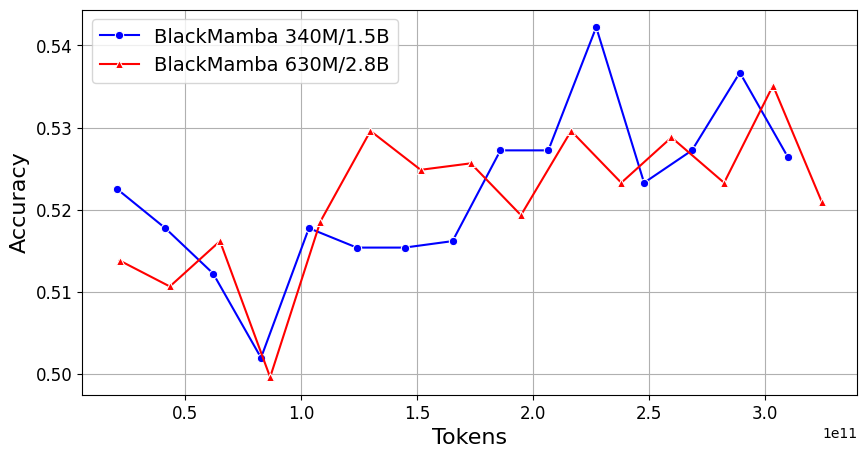}
    \caption{WinoGrande evaluation accuracy over time}
    \label{fig:winogrande}
\end{figure}

\begin{figure}[H]
\centering
    \includegraphics[width=.8\linewidth]{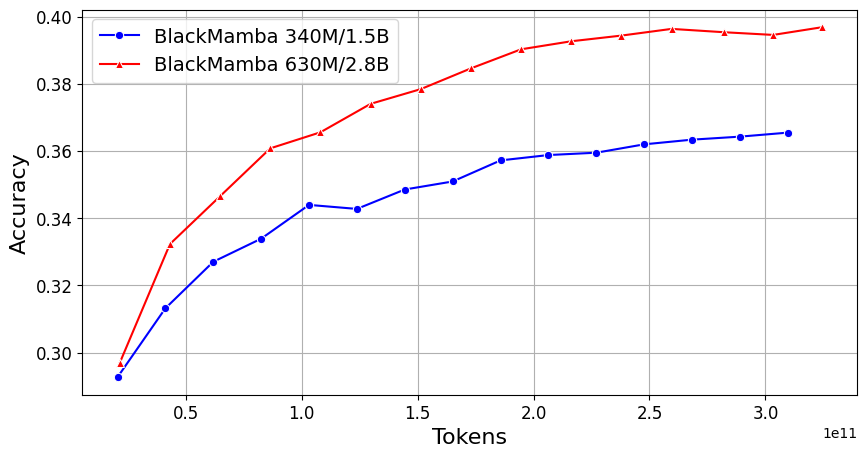}
    \caption{HellaSwag evaluation accuracy over time}
    \label{fig:hellaswag}
\end{figure}

\begin{figure}[H]
\centering
    \includegraphics[width=.8\linewidth]{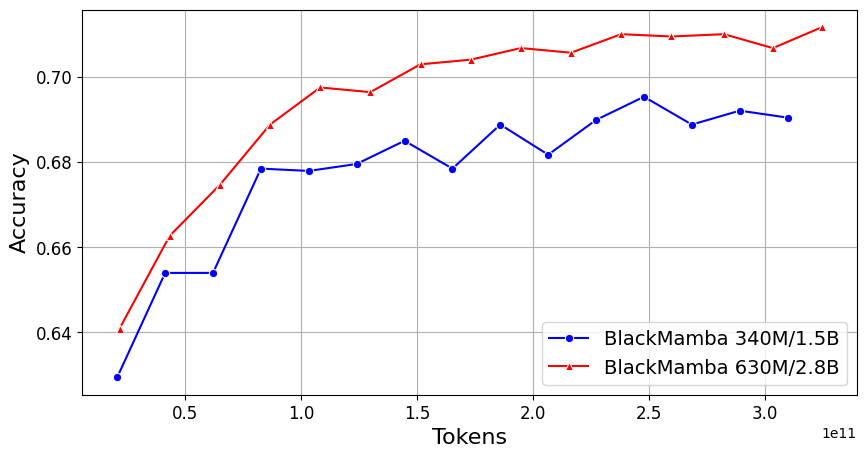}
    \caption{PIQA evaluation accuracy over time}
    \label{fig:piqa}
\end{figure}

\begin{figure}[H]
\centering
    \includegraphics[width=.8\linewidth]{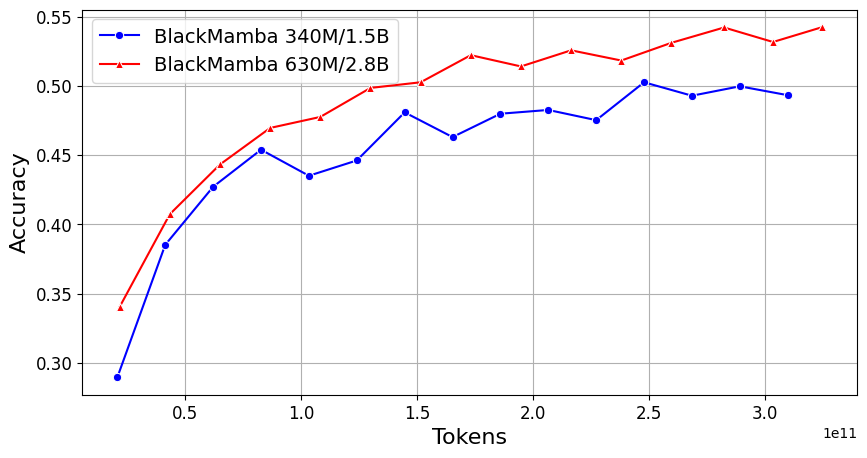}
    \caption{Lambada evaluation accuracy over time}
    \label{fig:lambada}
\end{figure}


\subsection{Sinkhorn MoE Routing Modifications}
\label{sec:appendix-sinkhorn}

Recall from the main text eq. (\ref{eq:moe}) that the output token $y$ of an MoE layer is given by
\begin{equation} y = \sum_{i\in \text{top-}k} c_i E_i(x)\end{equation}
where $E_1,E_2,\dots, E_N$ denote the MLP experts according to the top-$k$ probabilities $p_i$.

Most commonly, the probabilities $p_i(x)$ are obtained acting by a trainable linear layer on the input $x\in \mathbb R^d$ and subsequently applying a non-linearity: $p_i(x) = \sigma(W_i\cdot x)$, with $W_i\in \mathbb R^d$. An important issue when training MoE models is that expert utilization should be balanced across tokens in a batch, which is required for compute efficiency. Standard approaches to ensure balanced usage include adding a balancing regularization term to the loss as well imposing hard constraints bounding the number of tokens a given expert can receive \cite{lepikhin2020gshard}. We instead use the Sinkhorn activation function for the router which, in the context of top-1 expert selection, has proven to solve the balancing issue without the need for additional regularization or constraints on expert usage \cite{clark2022unified}. 

The key property of the Sinkhorn activation function is that, in addition to requiring normalization with respect to the expert index $i$ in $p_i(x)$, one additionally imposes normalization along the samples dimension (which comprises batch size and sequence length). More explicitly, we require that $\sigma$ satisfies:
\begin{equation} \label{eq:norm}\sum_{i=1}^N \sigma(W_i \cdot x_\alpha) = 1, \qquad \sum_{\alpha=1}^S \sigma(W_i \cdot x_\alpha) = S/N\end{equation}
where $\alpha$ denotes the sample index, and $S$ is the number of samples (batch size $\times$ sequence length). Now, note that the softmax, which only satisfies the first condition, can be variationally defined by maximizing:
\begin{equation}\label{eq:max} \text{softmax}(L) \equiv \text{argmax}_\pi\{ \pi\cdot L + S(\pi)\}\end{equation}
where $L_{i\alpha} = W_i\cdot x_\alpha$ are the logits, and $S(\pi)=-\sum_{i\alpha}\pi_{i\alpha}\log\pi_{i\alpha}$ is the Shannon entropy. The Sinkhorn activation can be defined through the same variational formulation except that it further satisfies the second constraint in (\ref{eq:norm}). Denoting the solution to this maximization by
\begin{equation} \pi_{i\alpha} = e^{L_{i\alpha}} d_i^{(0)}d^{(1)}_\alpha\end{equation}
where $d^{(0)}\in \mathbb R^N$ and $d^{(1)}\in \mathbb R^S$,
maximization of the right-hand side of (\ref{eq:max}) subject to (\ref{eq:norm}) is obtained by solving
\begin{equation} d^{(0)}_i = \frac 1{\sum_\alpha e^{L_{i\alpha}}d^{(1)}_\alpha},\qquad 
d^{(1)}_\alpha = \frac SN \frac 1{\sum_i e^{L_{i\alpha}}d^{(0)}_i}\end{equation}
Unfortunately, these equations cannot be solved explicitly and thus, unlike the softmax case, there is no analytic form for the Sinkhorn activation. These equations are solved approximately through an optimization loop, called the Sinkhorn algorithm \cite{sinkhorn1967concerning}.\footnote{We need to additionally choose $c_i$. One natural choice is $c_i=p_i$, but with the Sinkhorn activation we verified that it is more efficient to choose $c_i=f(W_i\cdot x)$ with $f$ a simple activation function such as the sigmoid. We think this is due to the Sinkhorn flattening out more quickly than e.g. sigmoid or softmax due to normalization along both dimensions.} Our improvement is in the choice of the initial condition for this optimization loop, which consists of taking $d^{(0)}_i=1$ and $d^{(1)}_\alpha = \frac SN \sum_i e^{L_{i\alpha}}$. This corresponds to initializing $\pi_{i\alpha}$ to be the softmax normalized along the sample index $\alpha$, thus immediately guaranteeing balanced usage of experts. We verified empirically that choosing this initial condition leads to much faster convergence of the Sinkhorn loop. Additionally, a temperature rescaling $L_{i\alpha}\to 2L_{i\alpha}$ further improves convergence. Overall this led to shrinking the number of iterations from 10-20 to just 1 across various models sizes, thus shortening the iteration time in our training experiments. 

\end{document}